\pdfoutput=1
\documentclass[11pt,a4paper]{article}

\usepackage[utf8]{inputenc}
\usepackage[T1]{fontenc}
\usepackage{lmodern}
\usepackage[margin=1in]{geometry}
\usepackage{amsmath,amssymb}
\usepackage{graphicx}
\usepackage{booktabs}
\usepackage{multirow}
\usepackage{xcolor}
\usepackage[hyphens]{url}
\usepackage[numbers,sort&compress]{natbib}
\usepackage{microtype}
\usepackage{tikz}
\usetikzlibrary{positioning,arrows.meta,fit,backgrounds}
\usepackage[colorlinks=true,linkcolor=blue!55!black,citecolor=blue!55!black,urlcolor=blue!55!black]{hyperref}

\usepackage{fancyhdr}
\begin{document}

\begin{center}
  {\LARGE\bfseries Interpretable EEG Microstates: Improving Modified K-Means Temporal Dynamics with a Recurrent, Person-Specific Variational Deep Embedding\par}
  \vspace{1em}
  {\large
    Saheed Faremi$^{1,2,\ast}$,\quad
    Andrea Visentin$^{1,3,\dagger}$,\quad
    Luca Longo$^{1,2,3,\dagger}$\par}
  \vspace{0.6em}
  {\small
    $^{1}$School of Computer Science and Information Technology, University College Cork, Western Road, Cork, T12 XF62, Ireland\\
    $^{2}$Artificial Intelligence and Cognitive Load Research Lab, University College Cork, Ireland\\
    $^{3}$Rinn Artificial Intelligence, University College Cork, Ireland\\
    \texttt{\{126100912@umail.ucc.ie, andrea.visentin@ucc.ie, luca.longo@ucc.ie\}}\\
    $^{\ast}$Corresponding author.\quad $^{\dagger}$All authors contributed equally.\par}
\end{center}
\vspace{0.5em}
\noindent\rule{\linewidth}{0.6pt}

\begin{abstract}
Electroencephalographic (EEG) microstates are the elementary units of
large-scale cortical dynamics, but the standard method, Modified K-Means,
returns only a hard label per frame and offers no representation to inspect.
A generative deep embedding, the Convolutional Variational Deep
Embedding model (Conv-VADE), learns microstate templates jointly with a
continuous latent space. The current study asks whether Conv-VADE reproduces the
established temporal dynamics of microstate sequences while providing an
organised, temporally structured latent representation that the geometric
baseline cannot offer. This is evaluated on the
eyes-closed resting-state EEG data from 203 LEMON participants against a matched
Modified K-Means baseline at cluster counts $K=3$ to $20$.
It combines a five-axis recurrence suite with three interpretability lenses: capacity,
internal dynamics and manifold faithfulness.
The model recovers almost equivalent maps, near-parity
variance explained after interpolation correction,
richer transition syntax and strong long-range correlations.
Its 32-dimensional latent space uses nearly all its capacity and is strongly recurrent beyond a phase-randomised null
(determinism $0.700$, $z=23.88$, 100\% of the 203 subjects). Microstate determinism
arises almost entirely from dwell time on observable axes in both methods. The latent
representation, not the label sequence, carries the learned temporal
structure, making generative deep embeddings an inspectable alternative to
geometric Modified K-Means microstate analysis.

\end{abstract}
\noindent\rule{\linewidth}{0.6pt}
\vspace{0.5em}

\section{Introduction}\label{sec1}
Microstates represent quasi-stable configurations of the scalp field. Empirical work suggests they last between 60 and 120 milliseconds. They are widely regarded as the basic building blocks of brain activity. Microstates serve as the atoms of large-scale cortical dynamics of the human brain \citep{michel2018eeg}. Several studies have analysed microstates and their segmentation. Lehmann's segmentation approach and the work of Pascual-Marqui et al.\ established Modified K-Means in microstate research \citep{pascualmarqui1995segmentation}. Geometric clustering of global-field-power (GFP) peak topographies is the standard of EEG analysis across subjects, paradigms and laboratories. It retrieves canonical maps, and each scalp field maps to exactly one microstate. However, it offers no representation beyond the label, and there exists no latent space, no generative replay, no calibrated uncertainty, and no principled way to query what a microstate is beyond its average scalp field.
The statistics derived from the label sequence (durations, coverage, and transition probabilities) form the basis of clinical and cognitive inferences in microstate research. A generative representation enables latent traversal, generative replay, continuous-trajectory recurrence analysis, and confidence assignment. These are capabilities that are unavailable to Modified K-Means because it returns only a discrete label.
To tackle these limitations, a Convolutional Variational Deep Embedding model (Conv-VADE) \citep{jiang2017variational} trained on interpolated scalp-field images is proposed.
It focuses on learning to generate topographies and clusters them via a Gaussian-mixture clustering prior over a low-dimensional latent space. Previous methods studies detail map recovery, variance, and pipeline validation \citep{faremi2026pilot,faremi2026brain}.
The proposed method grounds its acceptance in its capacity to replicate temporal dynamics observed in geometric segmentation and in the content of its latent representation being characterised.
The research question tackled in this research is: \emph{Can a generative deep embedding model, back-fitted to discrete microstate sequences, match the geometric baseline on the established dynamics of microstate analysis while providing an organised, temporally structured latent representation that the baseline cannot offer?}

The remainder of this article is organised as follows. Section~\ref{sec2} reviews related work on microstate segmentation, geometric clustering and generative approaches to EEG representation learning. The design and methodology are presented in Section~\ref{sec3}, which describes the LEMON dataset, the proposed Conv-VADE model, its back-fitting to discrete microstate sequences and the evaluation against the geometric baseline. Section~\ref{sec4} presents and discusses the experimental results on reproducing established microstate dynamics and providing an organised, temporally structured latent representation. Section~\ref{sec5} summarises the main conclusions and outlines directions for future work.

\section{Related Work}\label{sec2}
Lehmann's framework of quasi-stable field topographies underpins microstate analysis of multichannel EEG, and the canonical four-map solution has been recovered across laboratories and paradigms \citep{michel2018eeg}.
Map recovery of EEG signals has enabled the discretisation of non-stationary brain electrical activity, because the sequence isolates periods of quasi-stable activity, yielding accurate maps for cognitive workload, sleep stages, and consciousness analysis.
Recovering the sequence of these maps, however, requires a clustering method that analyses the topography and its temporal ordering.
The standard method used for these maps is Modified K-Means, which modifies the clustering method to account for EEG signals, with the introduction of polarity invariance using spatial correlations that allows the different poles of the brain signals to be treated as identical \citep{pascualmarqui1995segmentation,brunet2011spatiotemporal}.
The clustering algorithm assigns each time frame to the cluster with the highest absolute spatial correlation, and back-fitting the signal in this way yields the continuous label sequence.
The transition structure and dwell times are then computed from that sequence once the maps are fixed.
From the labels, the sequence is thereby described, not modelled.
To model rather than describe the sequence, hidden semi-Markov models define the probability distribution of state transitions and enforce temporal continuity through explicit-duration emissions \citep{vonwegner2017information}.
Coupling the labels from Modified K-Means with these probability distributions enables analysis of the duration assumptions underlying the label sequence.
However, in both approaches, the label sequence is modelled while the maps and the signal are given no generative representation.
Without a generative representation, internalisation of the maps and the signal is hard to test.
A generative model allows a test of whether the sequence structure reported under classical segmentation persists under generative segmentation.

Whether a generative representation reproduces the temporal structure of classical segmentation depends on what that structure is, and the descriptive literature has characterised it in two respects.
Firstly, dwell times are not memoryless, and correlations are long-range.
Von Wegner and colleagues tested whether the microstate label sequence goes beyond first-order structure on the basis of the dwell-time distribution, and the first-order model was rejected.
Dwell times thus retain information about past states, and Van De Ville et al.\ demonstrated that this memory extends across the whole sequence: temporal correlations in microstate labels follow a power law \citep{vandeville2010scalefree}.
Several findings have described the label sequence, but none considers the internal representation of the topographies that would generate it.
The notion that microstate maps and sequences are informative beyond the segmentation pipeline is independently supported by links to cortical generators \citep{custo2017electroencephalographic,pascualmarqui2014resting}, to frequency-specific information flow \citep{milz2017alpha}, to BOLD resting-state networks \citep{yuan2012spatiotemporal}, and to modalities of thinking \citep{milz2016functional}.
Additionally, cross-spectral matrices retain the generative information needed to recover the maps and their dynamics \citep{pascualmarqui2022crossspectra}.
The findings show that the temporal structure of the label sequence is documented in dwell times and correlations.
This supports the microstate phenomenon beyond the segmentation pipeline, and whether the dynamics survive generative segmentation is what no study has looked into or analysed.
Latent-state models link low-dimensional dynamics to cortical activity \citep{wang2023latent}, and learned representations have also been brought to microstate extraction itself \citep{chikkankod2023deep,chikkankod2024improving}.
Specifically, a unified representation-learning and clustering framework for two-dimensional topographic maps was proposed \citep{chikkankod2023deep}, and preliminary single-dataset gains over Modified K-Means were subsequently reported \citep{chikkankod2024improving}.
Another line of work was followed by Hu et al., who compared clustering strategies on naturalistic EEG \citep{hu2022microstate}. The above methods change how the maps are extracted, but they leave open how a learned model should be evaluated.

Empirical findings, especially on spectral data, point in two directions: F{\'e}rat et al.\ found frequency-band-dependent microstate dynamics \citep{ferat2022beyond}, whereas Croce et al.\ found that alpha-power associations were not necessarily band-specific \citep{croce2020alpha}, and post-encoding tasks alter occurrence and coverage \citep{dcrozbaron2021tasks}.
Together, these studies motivate spectral and temporal evaluation of generative segmentation, but they do not test recurrence in a learned latent trajectory.
Variational autoencoders \citep{kingma2014auto} have been applied across neuroscience, and the deep clustering extension VADE \citep{jiang2017variational} couples the encoder with a Gaussian-mixture prior to produce soft assignments jointly with the latent code.
Convolutional variational models have been trained directly on EEG topographic maps, their latent components linked to regional cortical activation \citep{ahmed2023latent}, and self-supervised convolutional models of spectral topographic head-maps learn quasi-stable high-level representations of cognitive activation \citep{longo2022brainrate}.
However, no convolutional VADE trained on interpolated scalp topographies has been validated against the documented microstate dynamics.
To tackle this, the method proposed in this research offers an architecture that follows the systematic search of the companion methods article \citep{faremi2026pilot} and the design reported in \citet{faremi2026brain}.
The interpretability of generative models is well studied, but standard evaluation toolkits rely on known ground-truth factors that resting-state EEG cannot provide. The beta-VAE \citep{higgins2017betavae} and its follow-up analyses \citep{burgess2018understanding,chen2018isolating} showed that objective reweighting encourages factorised latent dimensions.
Locatello and colleagues \citep{locatello2019challenging} then showed that unsupervised disentanglement is not identifiable without inductive biases, so the standard metrics (disentanglement, completeness, and informativeness [DCI], mutual information gap [MIG], separated attribute predictability [SAP], and modularity) \citep{eastwood2018framework,ridgeway2018learning} presuppose ground-truth factors of variation, and for EEG microstates no agreed factor catalogue exists.
Attribution tools, concept vectors \citep{kim2018interpretability}, Shapley values \citep{lundberg2017unified}, applied to EEG mental-workload models in \citet{raufi2024powershap}, and integrated gradients \citep{sundararajan2017axiomatic}, answer which input drives which output, not which latent dimension controls which topographic property.
Latent traversal provides the generative analogue by varying a single latent dimension and decoding the resulting points back into EEG topographic maps. As a consequence, interpretability for a microstate model has to proceed without ground-truth factors.
Recurrence quantification analysis \citep{marwan2007recurrence} measures the deterministic structure of a trajectory through diagonal-line statistics of the recurrence matrix, determinism (DET) being the most reported.
Determinism is the natural quantification of the non-random temporal organisation that the dwell-time and correlation studies above documented, since it counts recurrences that follow ordered, repeatable patterns rather than isolated coincidences.
The phase-randomised surrogate of Theiler and colleagues \citep{theiler1992surrogate} provides the null: the power spectrum is preserved while nonlinear structure is destroyed, so DET beyond the surrogate distribution indicates genuine temporal organisation.
Recurrence Quantification Analysis (RQA) has been applied to continuous EEG amplitude and, more rarely, to symbolic microstate sequences \citep{vonwegner2017information}.
Unfortunately, it has not been applied to the internal latent trajectory of a generative microstate model, nor systematically across a suite of representations of the same data.
In contrast, the present study applies RQA across five complementary representations of the same recordings (symbolic, symbolic-jump, continuous GFP, multivariate EEG, and the Conv-VADE latent trajectory) to compare the generative representation against Modified K-Means.

In synthesis, three gaps can be extracted from the literature above.
First, the dynamics of microstate sequences are well characterised, but whether they survive generative segmentation is untested.
Second, interpretability tools for generative models presuppose factors that EEG lacks.
Third, recurrence analysis has not been applied to a model's own internal representation.
A novel method, as presented in the next section, addresses all three: a five-axis recurrence suite tests the dynamics, and three factor-free lenses, namely capacity, latent recurrence, and generative artefacts, serve interpretability, all evaluated on the same trained models.

\section{Design and Methodology}\label{sec3}
The study tests the hypothesis: if a generative deep embedding,
trained with a convolutional encoder--decoder and a Gaussian-mixture
clustering prior over its latent space on the LEMON dataset, is evaluated
against Modified K-Means, then it matches the geometric baseline on maps,
corrected GEV, Markov syntax and long-range correlations, exceeds it in
symbolic-jump determinism, and yields a latent space that is used (most of
its capacity active, scaling with the number of states), recurrent beyond
the surrogate null and decodable to plausible topographies through its
generative artefacts.

\subsection{Data understanding and preparation}\label{subsec3a}
The study analyses eyes-closed resting-state EEG from the Leipzig Study for
Mind-Body-Emotion Interactions~\citep{babayan2019mind}.\footnote{LEMON:
\url{https://fcon_1000.projects.nitrc.org/indi/retro/MPI_LEMON.html}.}
The preprocessed dataset contains 203 participants with
56--61 scalp channels, sampled at 250\,Hz (100\,Hz for two participants). The
signal is re-referenced to the common average and band-pass filtered
(2--20\,Hz, zero-phase FIR). Global field power (GFP) is the across-channel
standard deviation, with peaks accepted at local maxima separated by at least
three samples (12\,ms at 250\,Hz; 30\,ms for the two 100\,Hz participants). Each peak's scalp topography (56--61 channels) is
projected to the scalp plane (azimuthal-equidistant) and interpolated with a
cubic spline to a $40\times40$ image, masked outside the electrode convex hull.
For each participant, topographies are split 90/10 by a fixed permutation
(seed~42) for training and early-stop validation; training-split pixel means
and standard deviations standardise both splits (clipped to $[-5,5]$), while
post-fit back-fitting operates on the full continuous recording.

\subsection{Model architecture}\label{subsec3b}

\begin{figure}[htbp]
\centering
\resizebox{0.80\textwidth}{!}{%
\begin{tikzpicture}[
  x=1cm, y=1cm,
  zlab/.style={font=\Large\bfseries, text=black},
  subbox/.style={draw=black!75, fill=white, rounded corners=1.5pt, line width=0.8pt},
  box/.style={draw=black!85, fill=white, rounded corners=1.5pt, align=center,
              inner xsep=7pt, inner ysep=6pt, line width=0.85pt,
              font=\Large, text=black},
  head/.style={box, minimum width=1.80cm, inner ysep=5pt},
  zcirc/.style={circle, draw=black, fill=white, line width=1.1pt,
                minimum size=12.6mm, inner sep=0.5pt, font=\LARGE\bfseries, text=black},
  ncirc/.style={circle, draw=black!75, fill=black!5, line width=0.8pt,
                minimum size=9.4mm, inner sep=0.3pt, font=\Large, text=black},
  arr/.style={-{Latex[length=2.8mm, width=1.9mm]}, line width=0.9pt,
              black!90, rounded corners=2pt},
  arrg/.style={arr, black!65},
  lab/.style={font=\Large, text=black, align=center},
  sublab/.style={font=\large, text=black!78, align=center},
  chlab/.style={font=\large\bfseries, text=black!85},
  slabf/.style={line width=0.7pt, draw=black!80, line join=round},
]

\def\slab#1#2#3#4#5#6#7{%
  \draw[slabf, fill=#6]
    (#1, #2+0.5*#4) -- ++(0.14,0.12) -- ++(#3,0) -- ++(-0.14,-0.12) -- cycle;
  \draw[slabf, fill=#7]
    (#1+#3, #2+0.5*#4) -- ++(0.14,0.12) -- ++(0,-#4) -- ++(-0.14,-0.12) -- cycle;
  \draw[slabf, fill=#5]
    (#1, #2-0.5*#4) rectangle ++(#3, #4);}

\node[zlab, anchor=west] at (1.00,14.55) {Input Representation};

\node[inner sep=1pt] (sensor) at (2.68,12.02)
  {\includegraphics[width=1.78cm]{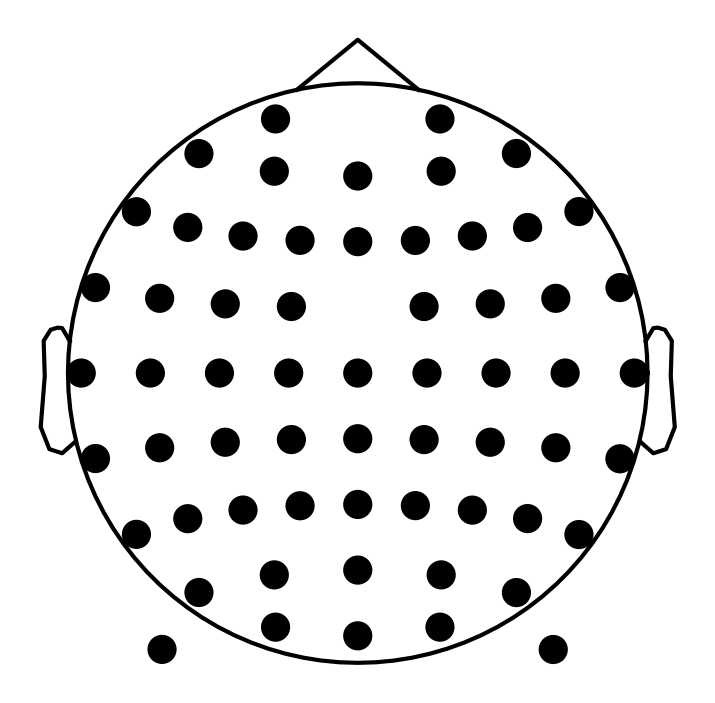}};
\node[lab] at (2.68,10.22)
  {EEG GFP Peak\\[-1pt]{\large\color{black!72}56--61 channels}};

\node[inner sep=1pt] (inputmap) at (6.38,12.02)
  {\includegraphics[width=1.78cm]{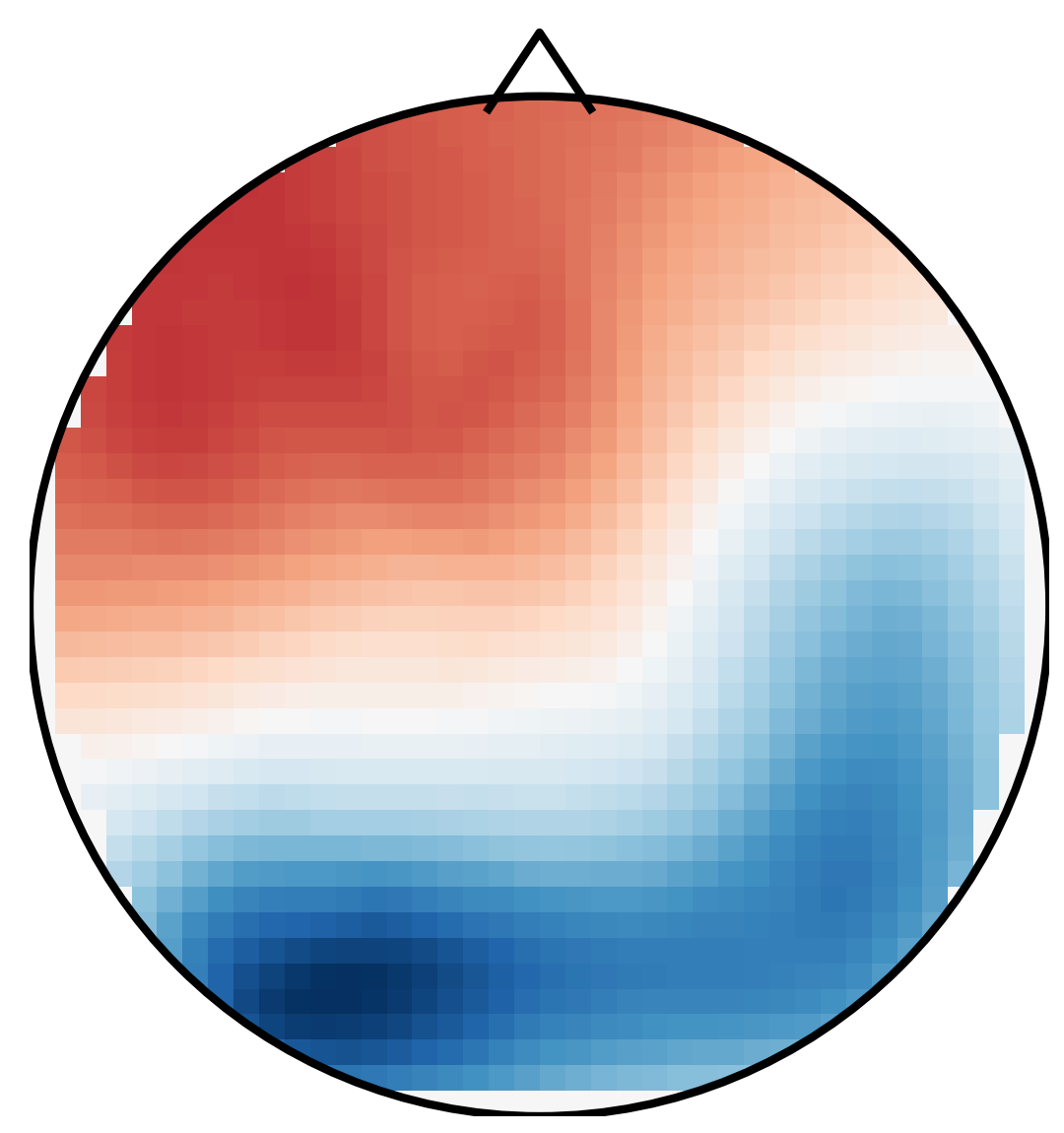}};
\node[lab] at (6.38,10.22)
  {Scalp Map $x$\\[-1pt]{\large\color{black!72}$40\times 40$ grid}};

\draw[arrg] (sensor.east) -- (inputmap.west);
\node[sublab] at (4.53,13.56)
  {Azimuthal equidistant\\[-1pt]projection \& spline};

\node[zlab, anchor=west] at (8.72,14.55)
  {Encoder $q_{\phi}(z \mid x)$};

\slab{8.74}{12.02}{0.46}{0.82}{black!10}{black!4}{black!22}
\slab{9.56}{12.02}{0.42}{1.14}{black!18}{black!8}{black!32}
\slab{10.36}{12.02}{0.36}{1.46}{black!27}{black!13}{black!42}
\slab{11.24}{12.02}{0.32}{1.78}{black!36}{black!18}{black!52}

\node[chlab] at (8.97,10.66) {32};
\node[chlab] at (9.77,10.66) {64};
\node[chlab] at (10.54,10.66) {128};
\node[chlab] at (11.40,10.66) {256};
\node[sublab] at (10.62,9.96) {$4\times$ Stride-2 Conv2D};

\node[box] (ops) at (15.32,12.02)
  {Global Avg.\ Pool\\[2pt]
   $1\times 1$ Conv (1024)\\[2pt]
   FC ($1024\to 512$)};

\draw[arr] (inputmap.east) -- (8.66,12.02);
\draw[arr] (11.80,12.02) -- (ops.west);

\node[zlab, anchor=west] at (1.00,8.35)
  {Latent Space ($d = 32$)};

\node[head] (mu) at (2.98,7.28) {$\mu$};
\node[head] (ls) at (2.98,5.88) {$\log\sigma^{2}$};
\node[ncirc] (eps) at (2.98,4.66) {$\varepsilon$};
\node[sublab] at (2.98,3.92) {$\varepsilon \sim \mathcal{N}(0, I)$};

\node[zcirc] (z) at (6.42,6.22) {$z$};
\node[sublab, text=black!85] at (6.06,7.62) {$z = \mu + \sigma \odot \varepsilon$};

\draw[line width=0.9pt, black!90, rounded corners=2pt]
  (ops.south) -- (15.32,9.15) -- (0.72,9.15) -- (0.72,6.58) -- (1.35,6.58);
\draw[arr] (1.35,6.58) |- (mu.west);
\draw[arr] (1.35,6.58) |- (ls.west);
\node[sublab, fill=white, inner sep=4pt] at (8.30,9.15)
  {512-dim.\ encoded representation};

\draw[arr] (mu.east) -- (z.158);
\draw[arr] (ls.east) -- (z.194);
\draw[arrg] (eps.east) -- (z.230);

\node[zlab, anchor=west] at (8.72,8.35)
  {Decoder $p_{\theta}(x \mid z)$};

\node[box] (dfc) at (9.94,6.22) {FC\\[1.5pt]$32\to 1024$};

\slab{12.08}{6.22}{0.32}{1.78}{black!36}{black!18}{black!52}
\slab{12.94}{6.22}{0.36}{1.46}{black!27}{black!13}{black!42}
\slab{13.72}{6.22}{0.42}{1.14}{black!18}{black!8}{black!32}
\slab{14.52}{6.22}{0.46}{0.82}{black!10}{black!4}{black!22}

\node[chlab] at (12.24,4.84) {256};
\node[chlab] at (13.12,4.84) {128};
\node[chlab] at (13.93,4.84) {64};
\node[chlab] at (14.75,4.84) {32};
\node[sublab] at (13.58,4.10) {$4\times$ Transposed Conv2D (symmetric)};

\node[box, inner xsep=6pt, inner ysep=6.5pt] (xhat) at (17.10,6.22)
  {$\hat{x}$\\[1.5pt]{\large $40\times 40$ grid}};

\draw[arr] (z.east) -- (dfc.west);
\draw[arr] (dfc.east) -- (12.00,6.22);
\draw[arr] (15.18,6.22) -- (xhat.west);

\node[zlab, anchor=west] at (1.08,2.52)
  {GMM Prior \& Assignment};

\draw[subbox] (1.08,0.00) rectangle (9.06,2.14);
\node[sublab, text=black!85] at (5.07,1.84)
  {\textbf{GMM Prior} ($K$ components)};

\draw[draw=black!55, densely dashed, fill=black!5, line width=0.6pt] (2.15,1.18) ellipse (0.58 and 0.28);
\draw[draw=black!80, fill=black!14, line width=0.65pt]                (2.15,1.18) ellipse (0.28 and 0.14);
\fill[black] (2.15,1.18) circle (0.045);

\draw[draw=black!55, densely dashed, fill=black!5, line width=0.6pt] (3.98,0.88) ellipse (0.50 and 0.24);
\draw[draw=black!80, fill=black!14, line width=0.65pt]                (3.98,0.88) ellipse (0.24 and 0.12);
\fill[black] (3.98,0.88) circle (0.045);

\draw[draw=black!55, densely dashed, fill=black!5, line width=0.6pt] (5.90,1.22) ellipse (0.62 and 0.27);
\draw[draw=black!80, fill=black!14, line width=0.65pt]                (5.90,1.22) ellipse (0.30 and 0.13);
\fill[black] (5.90,1.22) circle (0.045);

\draw[draw=black!55, densely dashed, fill=black!5, line width=0.6pt] (7.84,0.92) ellipse (0.50 and 0.24);
\draw[draw=black!80, fill=black!14, line width=0.65pt]                (7.84,0.92) ellipse (0.24 and 0.12);
\fill[black] (7.84,0.92) circle (0.045);

\node[lab] at (5.07,0.34)
  {$\{\pi_c,\ \mu_c,\ \log\sigma_c^{2}\}_{c=1}^{K}$};

\node[box, inner xsep=10pt, inner ysep=6pt] (g) at (5.07,-1.34)
  {$\gamma = q(c \mid z, x)$\\[2pt]
   {\large Microstate assignment}};

\draw[arrg] (z.south) -- (6.42,3.28) -| (8.52,2.14);
\draw[arr] (5.07,0.00) -- (g.north);

\node[zlab, anchor=west] at (10.14,2.52)
  {Microstate Template Extraction};

\node[inner sep=1pt] (centimg) at (14.25,1.08)
  {\includegraphics[width=4.9cm]{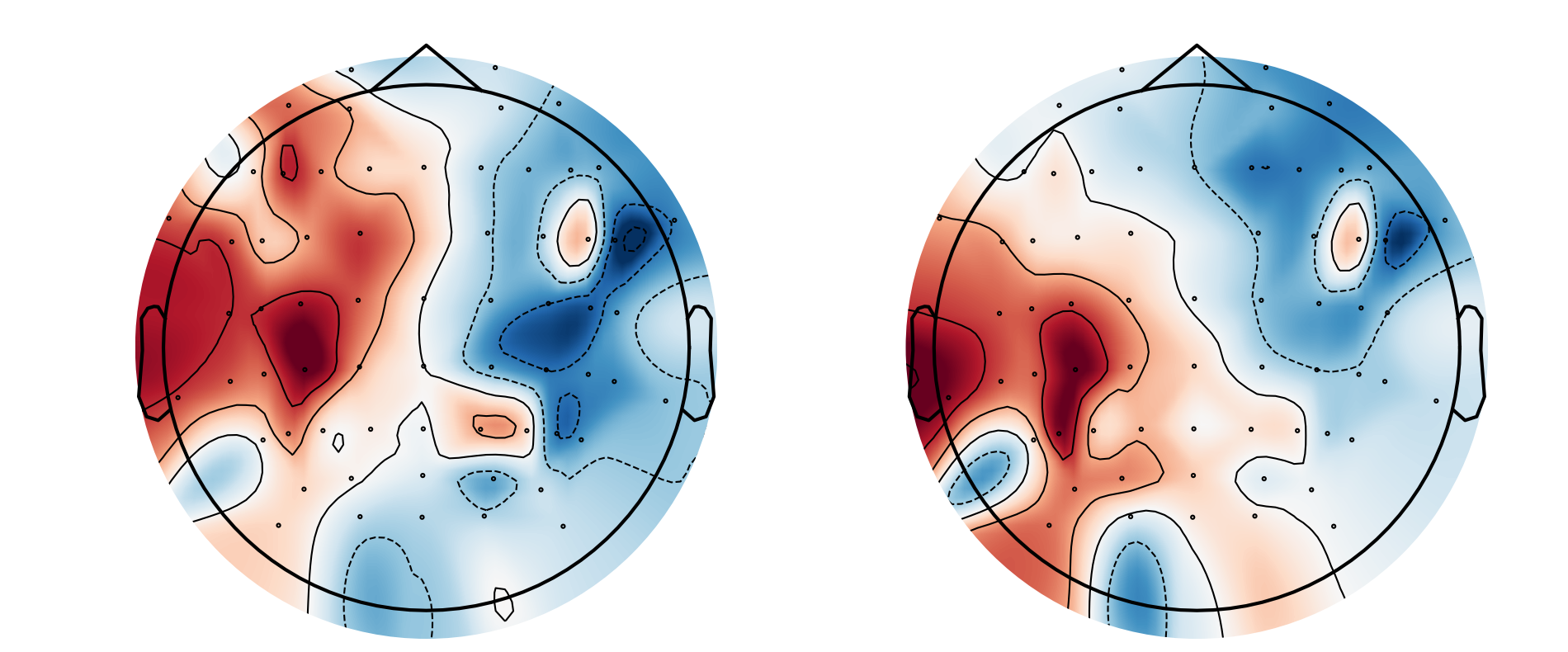}};
\node[lab] (tpllab) at (14.25,-1.14)
  {Centroid decoding $D(\mu_c)$\\[-0.5pt]
   \& grid-to-sensor resampling\\[2pt]
   {\large\color{black!72}$K$ scalp templates (56--61 channels)}};

\draw[arrg] (9.06,1.08) -- (centimg.west);
\node[sublab, fill=white, inner sep=2pt] at (9.68,1.52) {$\mu_c$};

\draw[arr] (tpllab.south) -- (14.25,-2.95) -- (0.00,-2.95)
           -- (0.00,12.02) -- (sensor.west);
\node[lab, fill=white, inner sep=5pt] at (7.12,-2.95)
  {Spatial back-fitting to sensor-space EEG};

\node[font=\Large, text=black, align=center] at (9.65,-4.08)
  {$\mathcal{L} = \mathcal{L}_{\mathrm{recon}}
   + \beta\,\mathcal{L}_{\mathrm{KL}}
   + 5.0\,\mathcal{L}_{\mathrm{b\text{-}ent}}
   + 0.2\,\mathcal{L}_{\mathrm{tight}}
   + 0.1\,\mathrm{MSE}\!\left(\mu(x),\,\mu(-x)\right)$};
\node[sublab] at (9.65,-4.82)
  {$\beta = 0.1$ ($10^{-3}$ pre-training) $\,\cdot\,$
   $\mathcal{L}_{\mathrm{recon}}$: reconstruction $\,\cdot\,$
   $\mathcal{L}_{\mathrm{KL}}$: GMM-prior KL};
\node[sublab] at (9.65,-5.42)
  {$\mathcal{L}_{\mathrm{b\text{-}ent}}$: entropy $\,\cdot\,$
   $\mathcal{L}_{\mathrm{tight}}$: tightening $\,\cdot\,$
   $\mathrm{MSE}$: polarity invariance};

\end{tikzpicture}
}
\caption{Convolutional VADE architecture for generative EEG microstate
segmentation. GFP-peak topographic maps ($40\times40$, interpolated from
56--61 channels) pass through four stride-2 convolutions to a
32-dimensional latent space ($\mu$, $\log\sigma^{2}$).}
\label{fig:arch}
\end{figure}

The model is configured over a shared architecture for
$K\in\{3,\dots,20\}$ with a $40\times40$ input (1{,}600 pixels). Four
stride-2 convolutional encoder blocks (channels $32\to64\to128\to256$), a
$1\times1$ bottleneck, and two fully connected layers produce $\mu$ and
$\log\sigma^2$, generating a 32-dimensional latent space with a
diagonal-covariance Gaussian-mixture prior yielding $q(c\mid z,x)$
\citep{jiang2017variational}. The decoder mirrors the encoder: a fully
connected layer expands $z$ back to 1{,}024 features, and four transposed
convolutions reconstruct the $40\times40$ topography. The polarity-invariant
objective, the KMeans initialisation of the mixture and the joint training
protocol follow the design validated on the same dataset in
\citet{faremi2026brain}, where a rank-based cluster-validity analysis gave a
modal $K=4$ for this model family.

\subsection{Model training and evaluation}\label{subsec3c}

\subsubsection{Model training}
Training minimises
\[
\mathcal{L}=\mathcal{L}_{\mathrm{recon}}+\beta\mathcal{L}_{\mathrm{KL}}
+w_{\mathrm{ent}}\mathcal{L}_{\mathrm{entropy}}
+w_{\mathrm{tight}}\mathcal{L}_{\mathrm{tighten}}
+w_{\mathrm{pol}}\mathcal{L}_{\mathrm{polarity}},
\]
where $\mathcal{L}_{\mathrm{polarity}}=\mathrm{MSE}(\mu(x),\mu(-x))$
enforces polarity invariance, with weights $w_{\mathrm{ent}}=5.0$,
$w_{\mathrm{tight}}=0.2$ and $w_{\mathrm{pol}}=0.1$.
Adam optimisation with learning rate $10^{-3}$ and batch size 128 trains
over 30 epochs of autoencoder pre-training and at most 100 epochs of main
training, with early stopping (patience 15) on the geometric mean of
silhouette and GEV, and $\beta=0.1$ during main training and $10^{-3}$
during pre-training. Each $K$ is trained independently.

\subsubsection{Model evaluation}
The matched baseline, Modified K-Means \citep{pascualmarqui1995segmentation},
is fitted on the same peaks with ten random initialisations, and the
highest-GEV run is kept. Due to the absence of in-sample GEV in Conv-VADE,
the comparison is post-fit to post-fit: decoded 56--61-channel centroids are
assigned to each GFP peak by highest absolute spatial correlation and
smoothed by the Pascual-Marqui procedure (factor 10, half-window 10, minimum
segment 8), using the same \texttt{pycrostates} routine as the baseline's prediction,
so differences in the post-fit sequences are attributable to the centroids. The observed transition matrix $T_{\mathrm{obs}}$ is estimated on the
repeat-collapsed sequence. The independence expectation is
$T_{\mathrm{exp}}(i,j)=\pi_j$, and the normalised deviation
$\Delta\mathrm{TM}_{ij}=
(T_{\mathrm{obs}}-T_{\mathrm{exp}})/T_{\mathrm{exp}}\times100\%$ marks
above- and below-chance transitions. Order-1 sufficiency is tested with a
likelihood-ratio comparison against a second-order chain. For each state's binary occupancy series, the Hurst exponent is estimated
using detrended fluctuation analysis \citep{peng1995quantification}. The
integrated series is centred on the mean and windowed at 12 log-spaced
scales, locally detrended, and the slope of $\log F(n)$ against $\log n$
gives $H$; $H>0.5$ indicates persistence.

RQA \citep{marwan2007recurrence} is applied to five representations: the
symbolic axis (dwell included), the symbolic-jump axis (dwell stripped), the
continuous-GFP axis (delay embedding $m=3$, $\tau=10$), the latent axis (the
32-dimensional trajectory), and the multivariate axis (raw 56--61-channel
vectors). Determinism is the fraction of recurrence points on diagonal lines
of length at least $l_{\min}=2$ (Theiler window 1). For continuous axes, distance thresholds are set per window to a fixed
recurrence rate of 0.05; for symbolic axes, recurrence is defined by exact
state equality.
DET is tested against 99 surrogates, phase-randomised for continuous axes
and first-order Markov for symbolic axes, with $z>2$ flagging recurrence
beyond the null. The computation is windowed into five windows of 1{,}000
samples. Shannon entropy, entropy rate, excess entropy and normalised Lempel--Ziv
complexity follow \citet{vonwegner2018information}. The analysis is repeated
on delta (1--4), theta (4--8), alpha (8--13) and beta (13--30\,Hz) bands by
band-pass filtering and back-fitting the broadband-trained centroids onto
band-limited peaks, so band differences reflect frequency content.

The active dimensions are analysed and counted. A latent dimension is active
if its across-sample posterior-mean variance exceeds $1\%$ of the per-run maximum. The active count per run
measures how much of the 32-dimensional capacity the model uses and whether
usage scales with $K$. Latent recurrence is read from the latent axis of the
RQA suite. Latent traversal interpolates one active dimension over its observed range, decoding each
point to a topographic map to expose what the dimension controls, while
generative replay samples the mixture prior and decodes, testing whether
sampled topographies resemble the fitted templates. Both artefacts are
bounded by design: a traversal shows what a dimension controls, not that it
is a named physiological factor, and generative replay shows local
faithfulness but not completeness.

All inferential statistics use subject-level aggregation: each subject's
metric is averaged across $K$ and then tested; $N=203$ for all metric
families (except active-dimension counts, exported for $N=143$ subjects
across 2{,}401 runs), avoiding pseudoreplication from the 18 repeated $K$
values per subject. The cross-method post-fit contrasts (Conv-VADE against
Modified K-Means) are complete at $N=203$. Wilcoxon tests are used for paired
comparisons and one-sample $t$-tests against nulls; 95\% confidence intervals
are mean $\pm1.96\,\mathrm{SE}$, Cohen's $d$ is reported for paired
comparisons, and all $p$-values are two-tailed.

\section{Results and Discussion}\label{sec4}
Hungarian-aligned absolute spatial correlation between Conv-VADE and
Modified K-Means centroids is $|r|=0.831$ (95\% CI $[0.828,0.834]$, $N=203$). This confirms that the
generative model recovers the canonical maps at every $K$. On variance
explained the comparison requires the interpolation correction, because the
decoded centroids carry a grid-to-channel attenuation that squared-correlation
GEV penalises. The raw post-fit GEV is $0.553$ against $0.608$ for Modified
K-Means (gap $0.055$, Wilcoxon $p=4.6\times10^{-35}$, $N=203$).
A single one-step template refinement, applied identically to both
methods and scored as the GFP-weighted mean squared correlation without
smoothing, yields $0.801$ for Conv-VADE against $0.821$ (unrounded gap $0.019$,
Wilcoxon $p=4.6\times10^{-35}$, $N=203$). Much of the apparent gap reflects the
interpolation attenuation rather than a representation deficit. After
correction the model approaches the geometric optimum; the small residual
reflects the cost of maintaining a generative latent space.

Symbolic RQA determinism is near-perfect for both Conv-VADE
($\mathrm{DET}=0.998\pm0.0005$, $z=1.52$, $11.7\%$ of runs recurrent) and
Modified K-Means ($\mathrm{DET}=0.997\pm0.0004$, $z=1.64$, $15.7\%$
recurrent; paired $\Delta=0.00044$, $d=2.16$, $p=4.8\times10^{-35}$,
$N=203$): in both models, DET near $1.0$ is a dwell artefact explained
largely by a first-order Markov surrogate. With dwell stripped, Conv-VADE
DET drops to $0.341$ ($z=0.54$), a likelihood-ratio test indicates
first-order sufficiency for $76.6\%$ of runs, and the entropy rate rises
with $K$ from $0.180$ ($K=3$) to $0.381$ ($K=20$) (subject-level means:
$|\Delta\mathrm{TM}|=35.6\pm15.5\%$, Shannon entropy $2.95\pm0.14$, excess
entropy $2.64\pm0.12$, normalised LZC $0.098\pm0.009$).
Microstate \emph{occupancy} is highly deterministic while the transition
\emph{rule} is near-random, reproducing the dwell-versus-switch dissociation
under generative segmentation.

On the symbolic-jump axis, which isolates transition syntax from dwell,
Conv-VADE exceeds Modified K-Means in both raw DET ($0.341\pm0.036$ vs.\
$0.267\pm0.013$; paired $\Delta=0.074$, Cohen's $d=2.66$,
$p=4.6\times10^{-35}$, $N=203$, all 203 subjects in the same direction) and
Markov-surrogate $z$-score ($+0.54$ vs.\ $-0.10$; paired $\Delta z=0.63$,
$d=0.68$, $p=2.2\times10^{-18}$, 166/203 subjects higher). Both models
remain close to first-order Markov behaviour in absolute terms, but
Conv-VADE produces consistently more structured transitions.
The Hurst exponent is
$H=0.833\pm0.022$ (100\% of 203 subjects above 0.5;
$t(202)=219.3$), indicating strong long-range persistence. $H$ correlates
positively with Conv-VADE GEV ($r=0.590$, $p=1.9\times10^{-20}$, $N=203$): subjects
whose maps explain more variance switch less often.

\begin{figure}[htbp]
\centering
\includegraphics[width=0.90\textwidth]{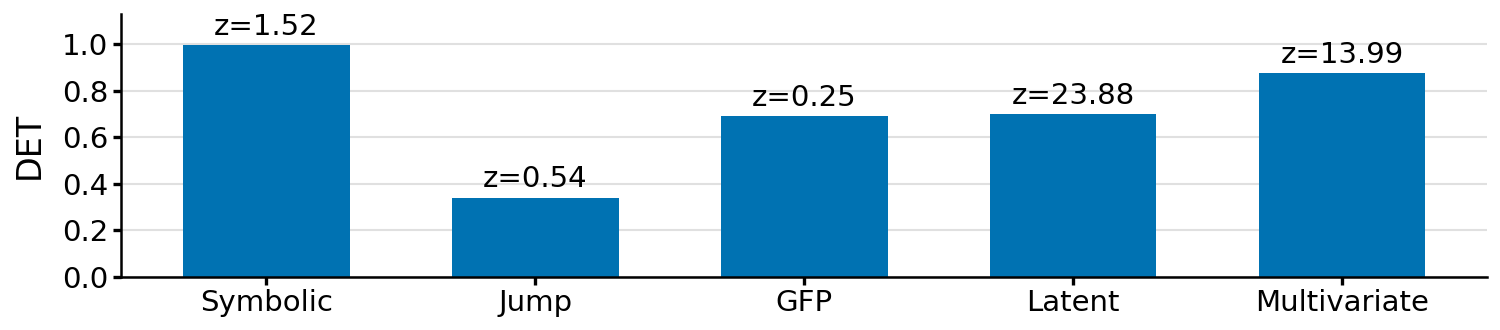}
\caption{Determinism across representations; labels show surrogate $z$-scores (cf.\ Table~\ref{tab:rqa}).}\label{fig:recurrence}
\end{figure}

The five RQA axes expose the dissociation no single axis can show
(Table~\ref{tab:rqa}). The symbolic axis is inflated by dwell and mostly
Markov-explained; the jump axis is near-random; the GFP axis does not recur
beyond phase noise ($z=0.25$); the multivariate axis is strongly recurrent
($\mathrm{DET}=0.875$, $z=13.99$); and the latent trajectory shows the
strongest recurrence relative to its surrogate null ($\mathrm{DET}=0.700$,
$z=23.88$), beyond the null in all 203 subjects (Figure~\ref{fig:recurrence}). The latent trajectory therefore retains
temporal structure that back-fitting discretisation discards.

\begin{table}[htbp]
\centering
\caption{Subject-level determinism ($\mathrm{DET}$, mean $\pm$ SD, $N=203$),
surrogate $z$-scores, and proportion of recurrent runs ($z>2$) across the
five RQA axes for Conv-VADE (VADE) and Modified K-Means (ModK).}\label{tab:rqa}
\small
\setlength{\tabcolsep}{4.5pt}
\begin{tabular}{@{}lcccccc@{}}
\toprule
& \multicolumn{2}{c}{DET} & \multicolumn{2}{c}{$z$ vs.\ surr.} & \multicolumn{2}{c}{\% rec.\ (runs)} \\
\cmidrule(lr){2-3}\cmidrule(lr){4-5}\cmidrule(lr){6-7}
Axis & VADE & ModK & VADE & ModK & VADE & ModK \\
\midrule
\multicolumn{7}{@{}l}{\emph{Model-dependent representations}} \\
Symbolic (dwell)  & $0.998\pm0.0005$ & $0.997\pm0.0004^{\ast}$ & $1.52$  & $1.64$  & $11.7\%$ & $15.7\%$ \\
Symbolic-jump     & $0.341\pm0.036$  & $0.267\pm0.013^{\dagger}$ & $0.54$  & $-0.10$ & $5.9\%$  & $9.1\%$  \\
Latent trajectory & $0.700\pm0.064$  & N/A$^{\ddagger}$        & $23.88$ & N/A     & $100\%$  & N/A      \\
\midrule
\multicolumn{7}{@{}l}{\emph{Signal-level reference representations (shared 56--61-ch EEG input)}} \\
Continuous-GFP    & \multicolumn{2}{c}{$0.692\pm0.080$} & \multicolumn{2}{c}{$0.25$}  & \multicolumn{2}{c}{$16.7\%$} \\
Multivariate full & \multicolumn{2}{c}{$0.875\pm0.053$} & \multicolumn{2}{c}{$13.99$} & \multicolumn{2}{c}{$100\%$}  \\
\bottomrule
\end{tabular}

\smallskip
\begin{minipage}{\linewidth}
\footnotesize
$^{\ast}$Paired (VADE $-$ ModK): $\Delta=0.00044$, $d=2.16$, Wilcoxon
$p=4.8\times10^{-35}$ ($N=203$).
$^{\dagger}$Paired DET: $\Delta=0.074$, $d=2.66$,
$p=4.6\times10^{-35}$; paired $z$: $\Delta z=0.63$, $d=0.68$,
$p=2.2\times10^{-18}$.
$^{\ddagger}$ModK returns discrete labels only (no latent space). Latent RQA
covers $3{,}651/3{,}654$ runs across all $N=203$ subjects ($99.97\%$ of runs
and $100\%$ of subjects with mean $z>2$).
\end{minipage}
\end{table}

\begin{figure}[htbp]
\centering
\includegraphics[width=0.95\textwidth]{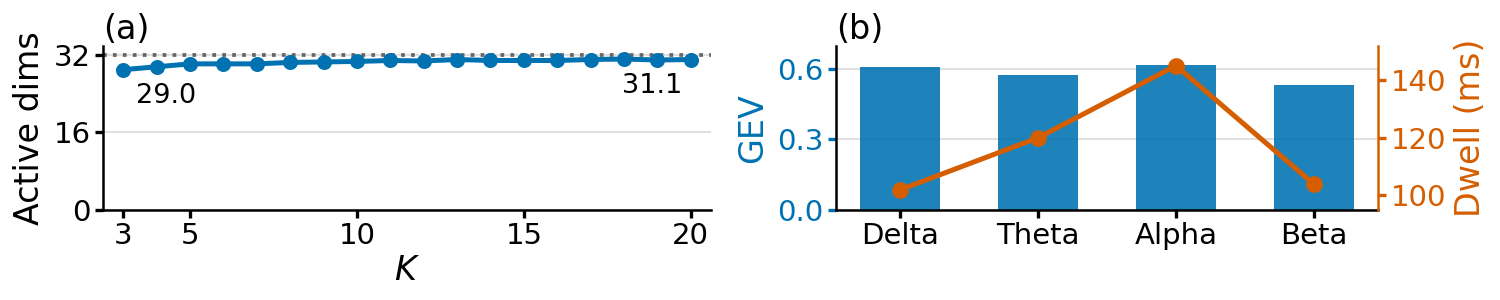}
\caption{(a) Active dimensions versus $K$; (b) band GEV (bars) and dwell duration (line).}\label{fig:active}
\end{figure}

Across runs with exported activity profiles ($N=143$, 2{,}401 runs), the
model uses $30.6$ of 32 latent dimensions on average, rising mildly with $K$
from $29.0$ ($K=3$) to $31.1$ ($K=20$) (Figure~\ref{fig:active}a).
Capacity scales with $K$ rather than collapsing, so
latent traversals remain meaningful.

Latent traversals produce smooth topographic deformations of the cluster
centroids as a single active dimension is interpolated, indicating that the
dimensions are organised rather than permutation-arbitrary. Samples from the
prior decode to topographies resembling the fitted centroids, providing a
qualitative check on the learned manifold. Together with the capacity and
recurrence results, these artefacts show that the latent space is
inspectable. The interpretation of both is bounded as set out in
Section~\ref{sec3}, and the per-run artefacts are released so the figures
are regenerable.

The same trained models allow a band-resolved read-out of the fitted
dynamics. Decomposing the label sequence by frequency band, using the
broadband-trained centroids, shows that the bands are not equivalent.
Alpha-band microstates (8--13\,Hz) explain the most variance
($\mathrm{GEV}=0.617$) and dwell longest ($145$\,ms); beta-band microstates
(13--30\,Hz) are fastest ($104$\,ms) and explain the least ($0.531$); and
delta (1--4\,Hz; $\mathrm{GEV}=0.609$, $102$\,ms) and theta (4--8\,Hz;
$\mathrm{GEV}=0.576$, $120$\,ms) fall between these extremes
(Figure~\ref{fig:active}b). The $8.6$-percentage-point ($\sim15\%$ relative) GEV spread across bands means
broadband statistics are alpha-weighted averages: two studies with different
alpha power will report different durations and coverage even for identical
generators.

The generative deep embedding reproduces properties of microstate dynamics, generates richer transition syntax than the geometric baseline, and yields a latent space that is usable, recurrent, and inspectable. The case for adoption is that it matches Modified K-Means on maps and nearly matches it on GEV, while representing microstates in a space where recurrence, traversal, and replay are tractable. Three limitations qualify these findings. First, the back-fit uses Pascual-Marqui smoothing, which inflates coverage relative to unsmoothed assignment. Second, latent recurrence is a model-internal statistic: it shows that the latent space is recurrent, not that this recurrence corresponds to an identifiable neural substrate. Third, counting active dimensions does not yield disentangled factors. Standard disentanglement metrics presuppose ground-truth factors for EEG, which do not exist \citep{locatello2019challenging}. Linking specific dimensions to physiological axes, such as anterior-posterior gradient, laterality, and band power, therefore remains an open problem.

\section{Conclusions and Future Work}\label{sec5}
The microstate field has standardised on geometric clustering methods that return only labels. Although several studies have characterised these labels, evaluating the interpretability of generative EEG models remains challenging because ground-truth factors are unavailable.
This study investigated whether a generative deep embedding could reproduce microstate sequence dynamics and examined what its latent space encodes.
Conv-VADE recovered equivalent maps ($|r|=0.831$) and achieved near-parity GEV after interpolation correction ($0.801$ vs.\ $0.821$).
It produced richer transitions than Modified K-Means (symbolic-jump DET $0.341$ vs.\ $0.267$, $d=2.66$; $z=+0.54$ vs.\ $-0.10$, $d=0.68$), preserved strong long-range correlations ($H=0.833$), and used nearly its full latent capacity ($30.6/32$).
Its central finding rests on a dissociation: microstate determinism on observable axes arose almost entirely from dwell time, whereas the model's latent trajectory was strongly recurrent in every subject tested (DET $0.700$, $z=23.88$, $N=203$).
These results indicate that discretisation discards temporal structure learned by the model, whereas the latent space retains it.
Future work will link active dimensions to physiological axes, extend the dwell-prior and smoothing audits to transition statistics, and replicate the methods with other datasets.

\section*{Acknowledgements}
This publication has emanated from research conducted with the financial support of Taighde \'Eireann -- Research Ireland under Grant No.~23/RC/13506. For the purpose of open access, the author has applied a CC BY public copyright licence to any Author Accepted Manuscript version arising from this submission.

\bibliographystyle{plainnat}
\bibliography{references}

\end{document}